\documentclass[twocolumn]{revtex4-1}
\usepackage[utf8]{inputenc}
\usepackage[english]{babel}

\usepackage{comment}
\usepackage{graphicx}
\usepackage{systeme,mathtools}
\usepackage{color}
\usepackage{xcolor,colortbl}
\usepackage{longtable}
\usepackage{csquotes}

\definecolor{Gray}{gray}{0.85}
\definecolor{LightCyan}{rgb}{0.88,1,1}

\newcolumntype{a}{>{\columncolor{Gray}}c}

\begin{document}

\title{Neural Network Training with Highly Incomplete Datasets}

\author{Yu-Wei Chang$^1$}\thanks{Equal contribution}

\author{Laura Natali$^1$}
\thanks{Equal contribution}

\author{Oveis Jamialahmadi$^2$}
\author{Stefano Romeo$^{\rm 2,3,4}$}
\author{Joana B. Pereira$^{\rm 5,6}$}
\author{Giovanni Volpe$^1$}
\thanks{Corresponding author giovanni.volpe@physics.gu.se}
\author{for the Alzheimer's Disease Neuroimaging Initiative}\thanks{Data used in preparation of this article were obtained from the Alzheimer’s Disease Neuroimaging Initiative (ADNI) database (\url{adni.loni.ucla.edu}). As such, the investigators within the
ADNI contributed to the design and implementation of ADNI and/or provided data but did not participate in analysis or writing of this report. A complete listing of ADNI investigators can be found at: \url{https://adni.loni.usc.edu/wp-content/uploads/how_to_apply/ADNI_Acknowledgment_List.pdf}.
}

\affiliation{$^1$Department of Physics, Gothenburg University, Gothenburg, Sweden}
\affiliation{$^2$Department of Molecular and Clinical Medicine, University of Gothenburg, Gothenburg, Sweden}
\affiliation{$^3$Clinical Nutrition Unit, Department of Medical
and Surgical Sciences, University Magna Graecia, Catanzaro, Italy}
\affiliation{$^4$Cardiology Department, Sahlgrenska University Hospital, Gothenburg, Sweden}
\affiliation{$^5$Department of Neurobiology, Care Sciences and Society, Karolinska Institutet, Stockholm, Sweden}
\affiliation{$^6$Clinical Memory Research Unit, Department of Clinical Sciences, Lund University, Malmö, Sweden}

\date{\today}

\begin{abstract}
Neural network training and validation rely on the availability of large high-quality datasets.
However, in many cases only incomplete datasets are available, particularly in health care  applications, where each patient typically undergoes different clinical procedures or can drop out of a study.
Since the data to train the neural networks need to be complete, most studies discard the incomplete datapoints, which reduces the size of the training data, or impute the missing features, which can lead to artefacts. 
Alas, both approaches are inadequate when a large portion of the data is missing.
Here, we introduce GapNet, an alternative deep-learning training approach that can use highly incomplete datasets. 
First, the dataset is split into subsets of samples containing all values for a certain cluster of features. Then, these subsets are used to train individual neural networks. Finally, this ensemble of neural networks is combined into a single neural network whose training is fine-tuned using all complete datapoints.
Using two highly incomplete real-world medical datasets, we show that GapNet improves the identification of patients with underlying Alzheimer’s disease pathology and of patients at risk of hospitalization due to Covid-19.
By distilling the information available in incomplete datasets without having to reduce their size or to impute missing values, GapNet will permit to extract valuable information from a wide range of datasets, benefiting diverse fields from medicine to engineering.
\end{abstract}

\maketitle

\section{Introduction}

Supervised machine-learning models, such as the neural networks employed in deep learning, require to be trained and validated on large high-quality datasets \cite{yanase_systematic_2019}.
In particular, these datasets need to be complete, i.e., each datapoint needs to have the values of all features, in order for them to be employed in standard neural-network training algorithms \cite{shilo_axes_2020}.
However, in many applications only incomplete datasets are available, i.e., each datapoint has values only for some of the relevant features \cite{little_prevention_2012}.
For example, this often occurs in medical applications involving patient data, e.g., because various patients might undergo different clinical and diagnostic procedures at different times, with some patients even dropping out from a study \cite{little_prevention_2012}.

In order to deal with these missing data, there are two standard approaches. 
The first and most commonly employed one is to exclude the datapoints that do not have all relevant features \cite{jakobsen2017}.
This approach has the advantage of ensuring the integrity of the data employed in training and validation, but it has the drawback of reducing the statistical power of the dataset \cite{ginkela2020} and, therefore, the predictive ability of the resulting deep-learning models \cite{sterne2009}.
Furthermore, data exclusion can introduce biases if the data are not missing completely at random \cite{kang2013}. 

The second and more complex approach is to impute the missing data. Various statistical imputation strategies have been proposed.
The simplest one is arguably to substitute the missing values with their ensemble average \cite{kang2013}.
More sophisticated imputation strategies obtain better results employing, e.g., multilayer perceptrons, extreme gradient boosting machines, and support vector machines \cite{liu_feature_2020, zhang_predicting_2020, huang_data_2016}.
For example, Vivar et al. \cite{vivar_simultaneous_2020} improved the classification of  individuals in the datasets of the Alzheimer’s Disease Neuroimaging Initiative (ADNI) and the Parkinson's Progression Markers Initiative (PPMI) by employing a multiple recurrent graph convolutional network to impute the missing features (the brain volumes obtained from magnetic resonance
imaging (MRI)). 
However, a major drawback of data imputation is that it can amplify biases in the available data or even introduce spurious correlations \cite{jakobsen2017}, especially when data are missing in great numbers or not at random  \cite{hughes2019}.
Because of their drawbacks, both exclusion and imputation strategies can only deal with a limited amount of missing data.

To address these limitations, here we introduce GapNet, an alternative neural-network training approach based on a hierarchical architecture that can make use of highly incomplete datasets.
GapNet takes advantage of all available information without the need for imputation of missing data.
First, the dataset is split into subsets containing all datapoints with a certain cluster of features. Then, these subsets are used to train individual neural networks. Finally, this ensemble of neural networks is combined into a single neural network whose training is fine-tuned using all complete datapoints.
As real-world test cases with highly incomplete datasets, we show that GapNet improves the identification of patients in the Alzheimer’s disease continuum in the ADNI cohort and of patients at risk of hospitalization due to Covid-19 in the UK BioBank cohort.
By distilling the information available in large incomplete datasets without having to reduce their size or to impute missing values, GapNet allows extracting valuable information from a wider range of datasets, being employable for many applications.

\begin{figure*}[t!]
    \centering
    \includegraphics[width=\linewidth]{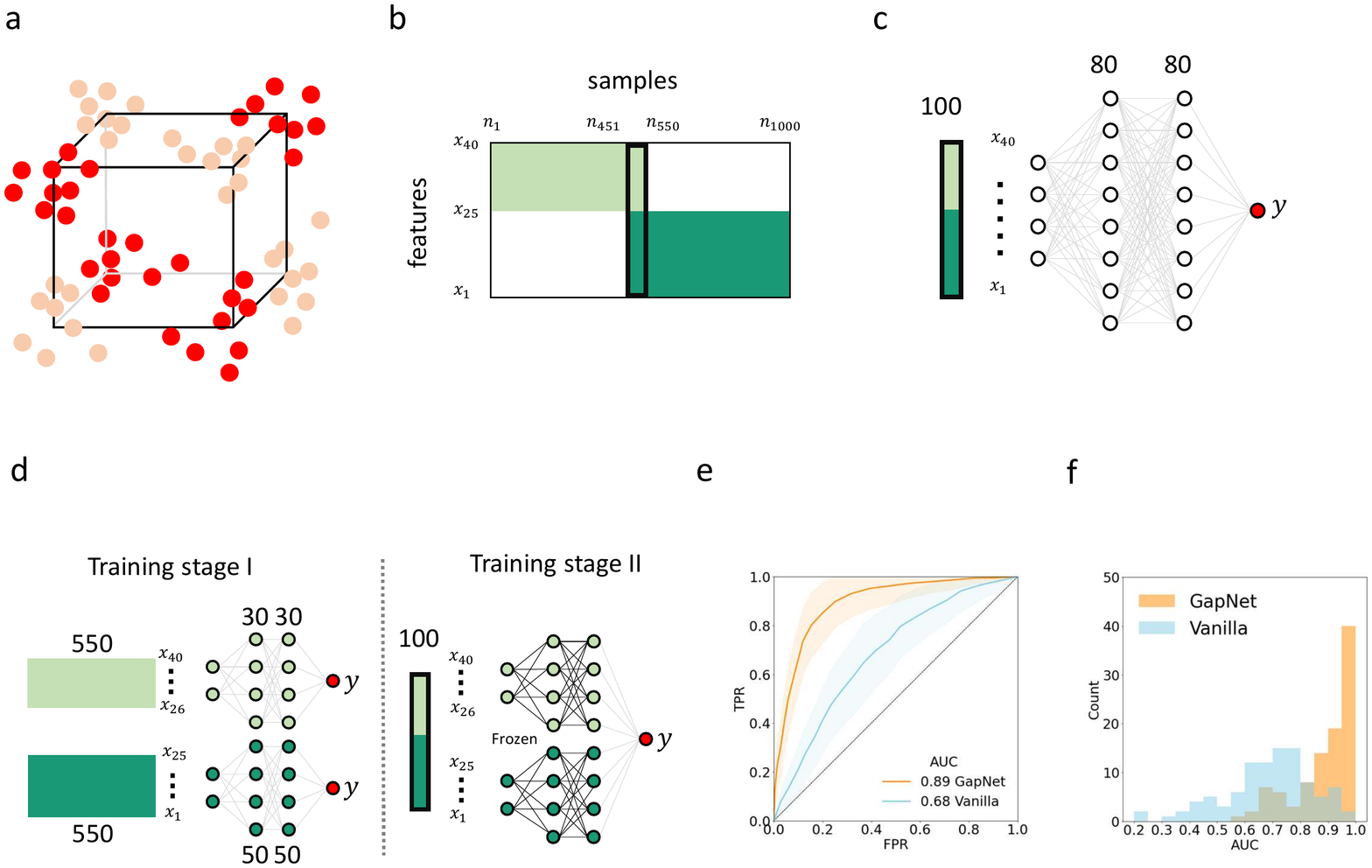}
    \caption{
    {\bf  GapNet improves the classification of simulated data.}
    {\bf a} Schematic representation of Madelon \cite{Guyon2005}, an artificial dataset used to test binary classification problems. The points in the dataset are clustered around the vertices of an hypercube with dimensions corresponding to the number of features ($F=3$ in the schematic) and attributed to $2$ different classes (corresponding to the pink and red circles in the schematic). 
    {\bf b} We simulate $N=1000$ samples with $F=40$ features ($x_1, ..., x_{40}$).
    Colored areas represent the available data, while the white areas represent the missing data: Only $100$ samples (samples $451$ to $550$, highlighted by the thick black line) have all $40$ features. 
    {\bf c} The baseline benchmark is provided by a vanilla neural network trained and tested on the complete samples.
    {\bf d} The GapNet approach: In training stage I, subsets containing all samples with certain features are used to train individual neural networks. In training stage II, their outputs are combined by an output node, whose connections are trained with all complete samples.
    {\bf e} The receiver operator characteristic (ROC) curve showing the true positive rate (TPR) versus the false positive rate (FPR) of the GapNet (orange, area under the curve (AUC) $0.89\pm0.11$) is significantly better than that of the vanilla neural network (blue, AUC $0.68\pm0.17$). Average (lines) and standard deviations (shaded areas) are obtained from 100 random splits of the training and testing datasets.
    The black dashed line represents the ROC curve of a random classifier. To establish statistical significance, the classifiers are also compared using the Delong test \cite{delong1988comparing} resulting in a significant p-value $<0.0001$ ($z=20.6$).
    {\bf f} Comparing the histograms of the AUC values for each independent run of the GapNet (orange) and the vanilla neural network (blue), we observe that GapNet delivers better results (larger AUC) in a more consistent way (smaller variance).
    }
    \label{fig1}
\end{figure*}

\begin{table*}[t!]
  \centering 
  \includegraphics[width=\linewidth]{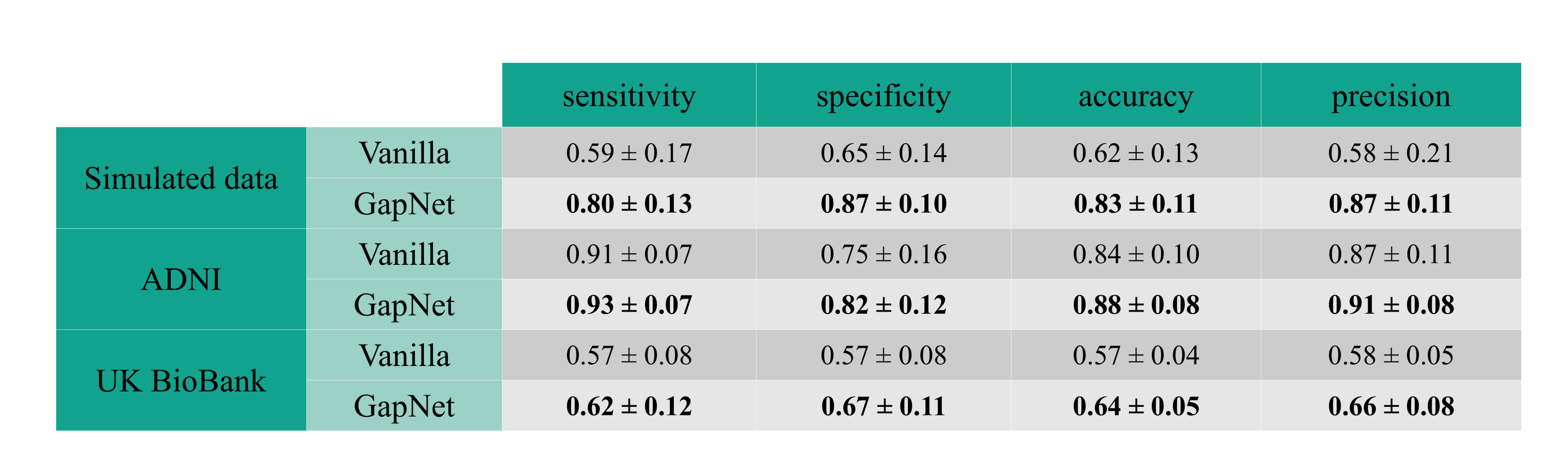}
    \caption{
    {\bf Performance of GapNet and vanilla neural network.}
    Comparison between the performance of GapNet and that of the vanilla neural network in terms of sensitivity, specificity, accuracy, and precision for the different datasets corresponding to simulated data (Figure~\ref{fig1}), neuroimaging data from the ADNI cohort (Figure~\ref{fig2}), and Covid-19 severity predictions from the UK BioBank cohort (Figure~\ref{fig3}). The numbers in bold represent the best results for each category. GapNet improves all the metrics for all datasets.
    }
    \label{tab1}
\end{table*}

\section{Results}

\subsection{GapNet working principle}

To demonstrate the GapNet working principle, we first apply it on a simulated dataset, representing a two-class classification problem with $F$ continuous input features. 
Specifically, we employ the simulated dataset Madelon \cite{Guyon2005}, where the datapoints are clustered around the vertices of an $F$-dimensional hypercube and assigned to the class of the closest vertex 
(see details in Methods, ``Simulated dataset''). 
Figure~\ref{fig1}a provides a schematic illustration for the case $F=3$.
Here, we consider a system with $F=40$ features, so that
\begin{equation}
    y = f(x_1, \dots, x_{40}),
\end{equation}
where $y \in \{0, 1\}$ is the class assignment for each datapoint.
As shown in Figure~\ref{fig1}b, the dataset consists of 1000 samples, where only 100 samples have all the 40 features: samples 1 to 450 have only features $x_{26}$ to $x_{40}$; samples 551 to 1000 have only features $x_{1}$ to $x_{25}$; while samples 451 to 550 have all features.

To establish a benchmark, we first consider a vanilla neural network approach, which is schematically shown in Figure~\ref{fig1}c.
We employ a dense neural network having an input layer with 40 nodes corresponding to the 40 features, two hidden layers with 80 nodes each (ReLU activation), a dropout layer with a frequency of 0.5, and an output layer with a single node (sigmoidal activation).
This vanilla neural network must be trained using the complete samples.
Thus, we split the available 100 complete samples into a 80-sample training set and a 20-sample testing set.
We train the neural network for 2000 epochs using the Adam optimizer \cite{kingma2014adam} with binary cross-entropy loss.

The GapNet approach involves a two-stage training process, as schematically presented in Figure~\ref{fig1}d. 
In training stage I (Figure~\ref{fig1}d), the input feature space is split into a set of non-overlapping clusters for which complete samples are available.
We then train a neural network using each of these clusters to predict the desired output.
In the current example, we identify two clusters of features, the first one corresponds to features 1 to 25 (dark green area in Figure~\ref{fig1}b) and the second one to features 26 to 40 (light green area), each with 550 samples.
Then, we train two dense neural networks to predict $y$ using $x_1, \dots, x_{25}$ and $x_{26}, \dots, x_{40}$, respectively.
Each neural network consists of an input layer (with 25 and 15 nodes, respectively, corresponding to each cluster of features), 
two hidden layers (with 50 and 30 nodes, respectively, with ReLU activation), a dropout layer (frequency 0.5), and finally an output layer with a single node (sigmoidal activation).
We train these neural networks on the available samples (retaining 20 complete samples for testing, the same for both neural networks) for 2000 epochs (Adam optimizer \cite{kingma2014adam}, binary cross-entropy loss).

In training stage II (Figure~\ref{fig1}d), the input and hidden layers of the first-stage neural networks are combined into a single neural network, adding an output node (sigmoidal activation).
These new connections are trained for 2000 epochs (Adam optimizer \cite{kingma2014adam}, binary cross-entropy loss) using all available complete samples (retaining for testing the same 20 samples used for testing in training stage I).

In order to statistically compare the performance of GapNet to the performance of the vanilla neural network, we repeat the training and testing procedures 100 times with resampling of the training and testing datasets.
We evaluate the performance using the  receiver operator characteristic (ROC) curve, which plots the true positive rate (TPR) versus the false positive rate (FPR) as a function of the threshold. 
The area under the ROC curve (AUC) is then 0.5 for a random estimator and approaches 1 as the estimator performance improves.
In Figure~\ref{fig1}e, we show the ROC curve for GapNet (orange line) and vanilla neural network (blue line), obtained as the average over 100 repetitions, while the corresponding variance is shown by the shaded areas.
The GapNet approach (AUC $0.89 \pm 0.11$) outperforms the vanilla neural network (AUC $0.68\pm0.17$). This difference is statistically significant with a p-value $p<0.0001$ ($z= 20.6$), tested using the Delong test \cite{delong1988comparing}.
In addition to the AUC  comparison, Table~\ref{tab1}  shows  the  sensitivity,  specificity,  accuracy,  and precision  computed  by  setting  the threshold at $0.5$.  All the metrics are improved by using GapNet, showing its ability to correctly identify the true positive and true negative cases.
Interestingly, the variability of the GapNet approach is smaller than that of the vanilla neural network approach, indicating more consistent training results.
Figure~\ref{fig1}f shows the histograms of the AUC values for each independent run. 
The peak for the GapNet is larger than that of the vanilla neural network, indicating GapNet achieves a consistently better performance, increasing the robustness of the classification to the missing data.

Overall, these proof-of-principle results on simulated data show the effectiveness and robustness of the GapNet approach to fully exploit incomplete datasets.
Thus, the clear advantage is that GapNet can make use of all available data, without having to impute the missing data.

\begin{figure*}[t!]
    \centering
    \includegraphics[width=\linewidth]{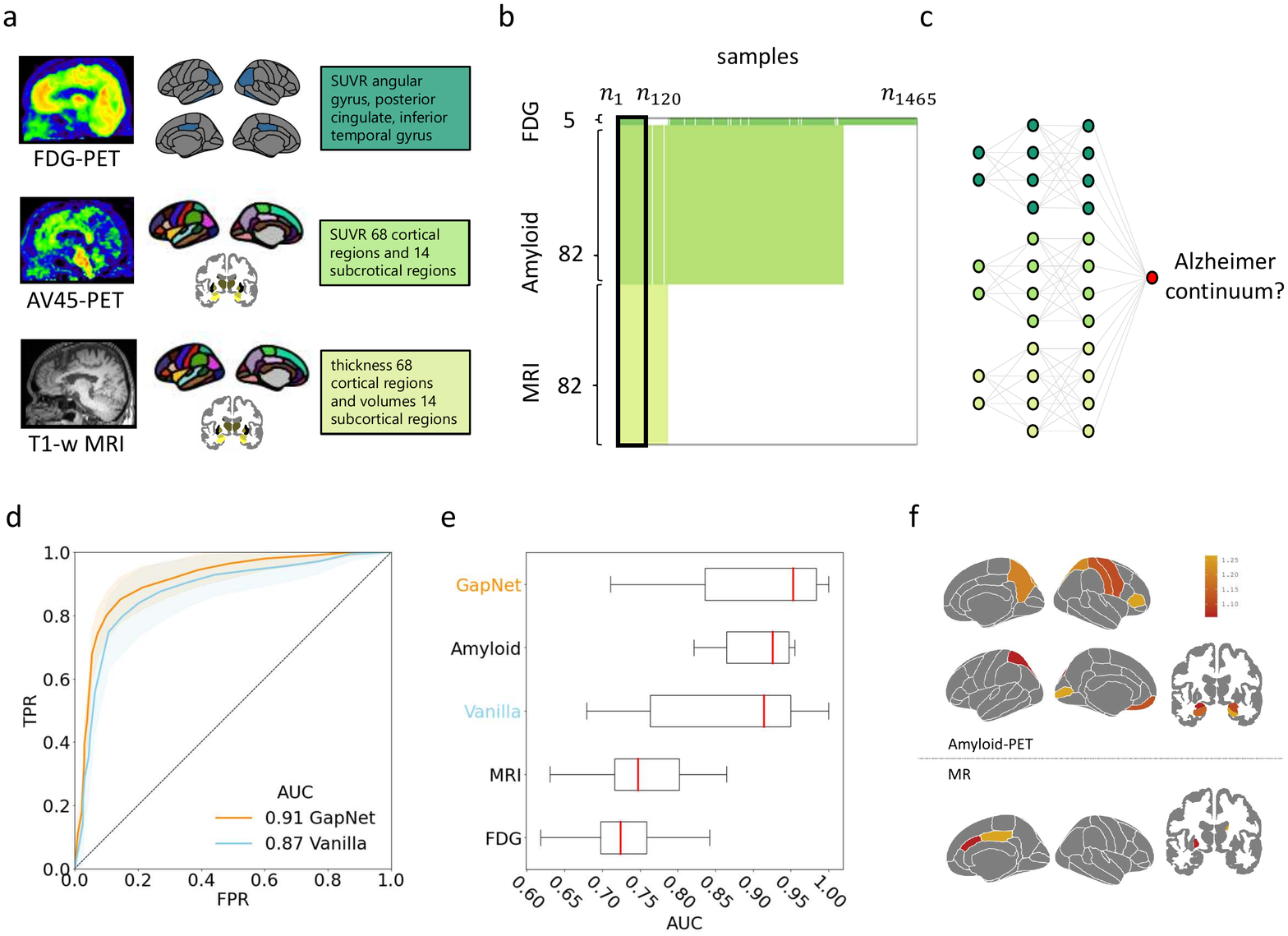}
    \caption{
    {\bf GapNet improves the identification of patients on the Alzheimer’s disease continuum in the ADNI cohort.}
    {\bf a} Neuroimaging modalities (structural magnetic resonance (MRI), amyloid positron emission tomography (amyloid-PET), and fluorodeoxyglucose-PET (FDG-PET)) and corresponding brain regions of interest (ROIs). Mean values in each ROI are extracted by a standard Freesurfer processing pipeline.
    {\bf b} Schematic representation of the dataset. 
    The dataset corresponds to $1465$ visits in each of which one or more neuroimaging modalities are employed.
    Only $120$ visits resulted in the acquisition of all three neuroimaging modalities (highlighted by the thick black line).
    {\bf c} GapNet: First a neural network with two hidden layers is trained using each cluster of features (i.e., the brain region data for each neuroimaging modality). Then, all these neural networks are concatenated to return the final output. 
    {\bf d} Receiver operating characteristic (ROC) curve showing that GapNet (orange, AUC $0.91\pm0.09$) outperforms the vanilla neural network (blue, AUC $0.87\pm0.11$). 
    The solid lines represent the averages over 100 independent runs and the shaded areas the corresponding standard deviations.
    The black dashed line represents the ROC curve of a random classifier. The classifiers are also compared using the Delong test \cite{delong1988comparing} resulting in a significant p-value $p<0.0001$ ($z=7.76$).
    {\bf e} Box plots of the AUC values over the independent runs showing that GapNet outperforms the vanilla neural networks trained either on all individuals with all diagnostic modalities (``Vanilla'') or on only one diagnostic modality (``Amyloid'', ``MRI'', or ``FDG'').
    The red mark represents the median, the black boxes are the interquartile ranges, and the black horizontal lines are the whiskers.
    {\bf f} Relevance heatmaps for the GapNet classification between cognitively normal individuals and patients in the Alzheimer's disease continuum. The color-coded areas represent the regions with high classification importance. For amyloid-PET (upper panel), these regions include the left pericalcarine, right pars trianularis, bilateral superior parietal lobule, right precuneus, right postcentral lobule, left medial orbitofrontal cortex, right precentral lobule at the cortical level, as well as bilateral hippocampus and bilateral amygdala at the subcortical level. For MRI (lower panel), the regions are mainly from the right hemisphere, including the caudal anterior cingulate and posterior cingulate at the cortical level as well as the right caudate and left pallidum at the subcortical level.
    }
    \label{fig2}
\end{figure*}

\subsection{Identification of patients in the Alzheimer’s disease continuum in the ADNI cohort}

As a first real-world GapNet application, we consider the identification of individuals with underlying amyloid pathology, which is one of the earliest pathological changes occurring in Alzheimer’s disease (AD) \cite{jack-2013, aizenstein-2008, lim-2013, vlassenko-2016}. 
We use data from the ADNI cohort
(see Methods, ``ADNI cohort''). 
In total, $869$ individuals underwent $1465$ neuroimaging visits including both baseline visits and subsequent longitudinal follow-up visits for some individuals.
In each visit, one or more of the following three neuroimaging modalities were employed: structural magnetic resonance (MRI), amyloid positron emission tomography (amyloid-PET),  and  fluorodeoxyglucose-PET (FDG-PET), which are normally used to assess gray matter atrophy, amyloid deposition and glucose hypometabolism in AD, respectively (Figure~\ref{fig2}a). 
For MRI, we include the mean thickness of the 68 cortical regions of the Desikan atlas \cite{desikan-2006} and the volumes of $14$ subcortical gray matter regions of the Aseg atlas \cite{fischl2002}.
For amyloid-PET, we include the mean amyloid standard uptake value ratio (SUVR) values from the same brain regions included in MRI. 
For FDG-PET, we include the SUVR of 5 composite brain regions \cite{landau-2011}.
The final number of features is $169$, and the corresponding regions of interest (ROIs) are listed in Supplementary Table~S1.
MRI scans were acquired in $233$ visits, amyloid-PET scans in $1045$ visits, and FDG-PET scans in $1258$. Only $120$ visits (corresponding to $118$ individuals, of which $40$ cognitively normal subjects without amyloid pathology and $78$ subjects with amyloid pathology) resulted in the acquisition of all three neuroimaging modalities (Figure~\ref{fig2}b).

To apply the GapNet approach (Figure~\ref{fig2}c), we identify three clusters of features corresponding to the three imaging modalities.
We split the data into a training set and a testing set. 
We define the testing set from $20\%$ of the complete data ($N_{\rm test}= 24$). 
We use the rest of the data for the training set ($N_{\rm train}= 1441$ including the remaining $80\%$ of the samples with complete data as well as all the incomplete samples).
A schematic of the GapNet architecture is shown in Figure~\ref{fig2}c.
In training stage I, these three clusters of features are used to train three independent neural networks (input layer with $5$, $82$ and $82$ nodes, respectively; two hidden layers with $10$, $164$, and $164$ nodes, respectively, with ReLu activation; dropout layer with frequency $0.5$; output layer with single neuron and sigmoidal activation).
We train the MRI network on $209$ samples, the amyloid-PET network on $1021$ samples, and the FDG-PET network on $1234$ samples (in all cases for 2000 epochs with binary cross-entropy loss function and Adam optimizer).
These three networks are then combined into a single neural network in training stage II with a joint output node, and the new connections are retrained on the 96 complete training samples (2000 epochs, binary cross-entropy loss function, Adam optimizer).

The ROC curves in Figure~\ref{fig2}d show that the classification performance of the GapNet approach (orange, AUC $0.91\pm0.09$) is superior than that of the standard vanilla neural network (blue, AUC $0.87\pm0.11$).
The shaded areas represent the variation of these ROC curves over 100 random splits between the training and testing sets.
This difference is statistically significant (Delong test \cite{delong1988comparing}, p-value $p<0.0001$, $z= 7.76$). 
In addition to the AUC comparison, Table~\ref{tab1} shows the sensitivity, specificity, accuracy and precision computed by setting the prediction threshold at $0.5$. 
All the metrics are improved by using GapNet, showing its ability to correctly identify the true positive and true negative cases \cite{trevethan2017}.

Figure~\ref{fig2}e shows the box plots with the AUC values obtained over the independent runs for the GapNet approach, the networks trained on each of the data clusters, and the vanilla neural network.
The GapNet approach achieves the best performance, outperforming the vanilla neural network as well as all networks trained on a single data cluster.
Interestingly, the amyloid-PET-trained neural network also outperforms the vanilla neural network approach, probably taking advantage of its larger training set ($1021$ instead of $94$ samples). 
Nevertheless, both the MRI-trained and the FDG-PET trained neural networks underperform the vanilla neural network, despite having access to more ($233$ and $1258$, respectively) samples.

In Figure~\ref{fig2}f, we mapped the feature importance for the GapNet model by performing a permutation feature analysis \cite{molnar2019}.
The results show that $16$ out of the most important $20$ features are derived from the amyloid-PET imaging modality, and the remaining $4$ features are from MRI.
For amyloid-PET, the features included occipital (left pericalcarine gyrus), frontal (right pars triangularis, right postcentral gryus, left medial orbitofrontal cortex, right precentral gyrus), parietal (bilateral superior parietal gyri, right precuneus), and 
subcortical areas (bilateral hippocampus, bilateral amygdala).
For MRI, the most important features included cortical thickness in limbic (right caudal anterior cingulate and right posterior cingulate) as well as volumes in subcortical areas (right caudate, left pallidum).
Crucially, most of these features (or ROIs) have been reported to be impaired in AD by previous studies assessing patients at different disease stages.
For amyloid-PET, the orbito-frontal and precuneus ROIs have been often identified in the early stages of amyloid pathology \cite{palmqvist-2016}; 
the subcortical ROIs (amygdala and hippocampus) are in line with the A$\beta$42 accumulation regions that have been reported in amyloid pathology during the early AD stages;
and the other three cortical ROIs (pericalcarine, postcentral, and precentral) are consistent with the A$\beta$42 accumulation regions showing high SUVRs during the late AD stages \cite{palmqvist-2016}.
For MRI, both the hightlighted cortical ROIs (caudal anterior cingulate and posterior cingulate) and subcortical ROIs (caudate and pallidum) have also been reported in previous studies on informative regions across different stages of the AD continuum \cite{kautzky-2018, jones-2005, fennema-notestine-2009, davatzikos-2011, madsen-2010, rallabandi-2020}.

\begin{figure*}[t!]
    \includegraphics[width=\linewidth]{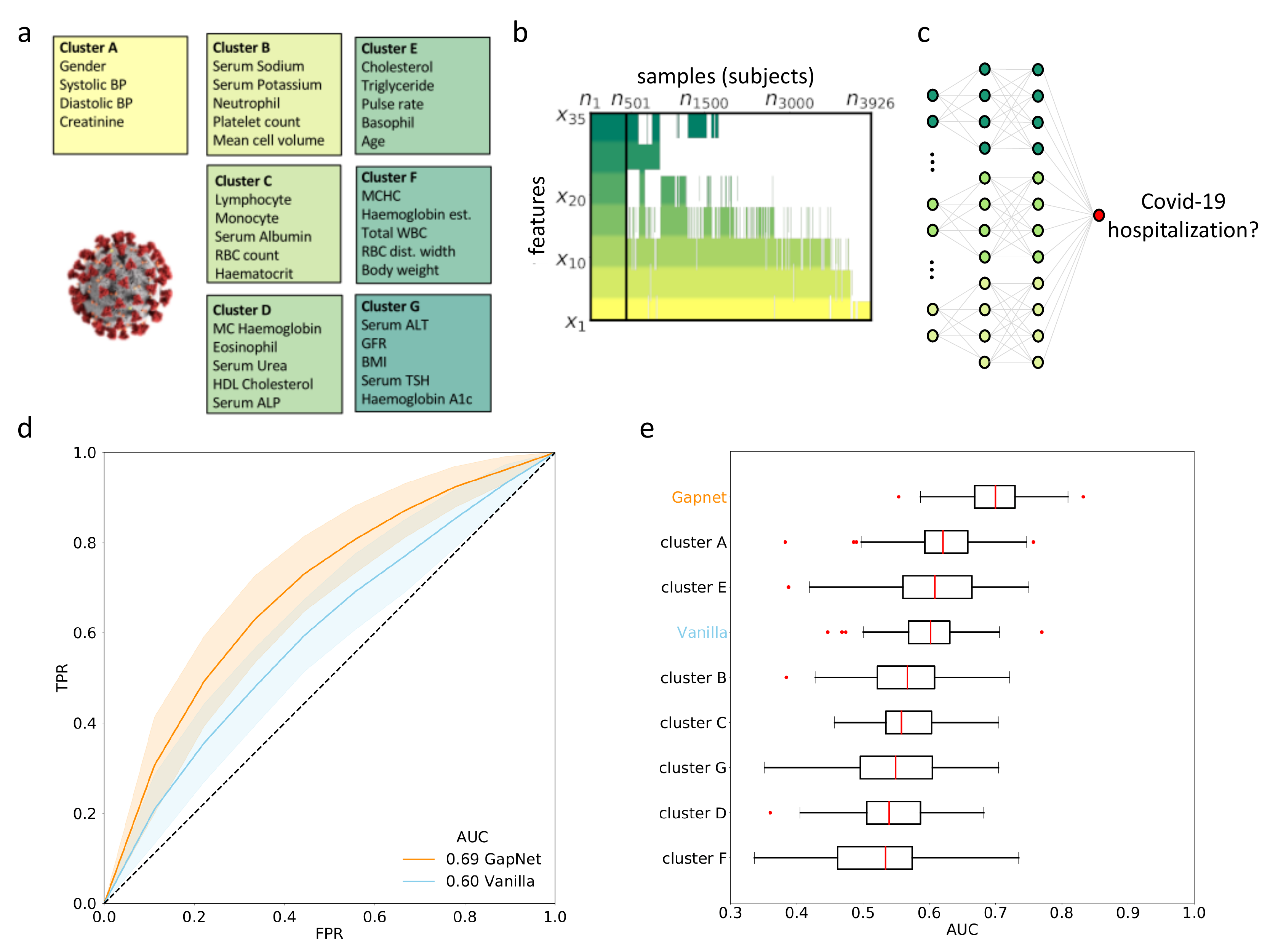}
    \caption{
    {\bf GapNet improves the identification of patients at risk of hospitalization due to Covid-19 in the UK BioBank cohort.}
    {\bf a} Seven clusters of features employed to predict hospitalization due to Covid-19 (sizes $A \rightarrow 3875$, $B \rightarrow 3713$, $C \rightarrow 3622$, $D \rightarrow 3109$, $E \rightarrow 1992$, $F \rightarrow 1123$, and $G \rightarrow 1536$).
    {\bf b} The dataset comprises $3926$ subjects and $34$ features with colors representing the different clusters from $A$ (yellow) to $G$ (dark green).
    The white areas represent the missing data.
    The $501$ subjects with all features are delimited by black lines. 
    {\bf c} Network architecture employed for the GapNet approach. 
    Neural networks with two hidden layers are trained for each cluster and then concatenated to obtain a single output. 
    {\bf d} Receiver operating characteristic (ROC) curve showing that GapNet (orange, AUC $0.69 \pm 0.07$) outperforms the vanilla neural network (blue, AUC $0.60 \pm 0.07$).
    The solid lines represent the averages over 100 independent runs and the shaded areas the corresponding standard deviations.
    The black dashed line represents the ROC curve of a random classifier.
    The classifiers are also compared using the Delong test \cite{delong1988comparing} resulting in a significant p-value $p<0.0001$ ($z=11.8$).
    {\bf e} Box plots showing the statistics of the AUC over the independent runs. 
    The red vertical lines mark the median value for each boxplot, the red dots are the fliers, the black boxes are the interquartile ranges, and the black horizontal lines are the whiskers.
    The performance of GapNet is better than that of the vanilla neural network.
    Furthermore, the results are compared with the training of neural networks on the single clusters and the plot shows them in descending order of medians (see the Supplementary Table~S2 for the complete list of features and the extended names).
    The GapNet approach features the best outcome. Interestingly, the features in clusters $A$ and $E$ alone result in better classifiers than the vanilla neural network.}
    \label{fig3}
\end{figure*}

\subsection{Prediction of Covid-19 hospitalization in the UK BioBank cohort}

As a second example of a real-world case application, we consider an incomplete dataset to predict hospitalization due to Covid-19. The dataset is based on the UK BioBank cohort, which contains information concerning Covid-19 test results, hospitalizations, and clinical examinations 
(see details in Methods, ``UK BioBank cohort''). 
The aim of this analysis is to discriminate patients at high risk of hospitalization due to severe Covid-19 symptoms from those at low risk of hospitalization, based on their previous medical records. 

The cohort includes $3926$ individuals and $34$ different features with a varying number of records per feature, ranging from $1123$ to $3875$ values. 
The parameters include sex, age and easily accessible testing information, such as red blood and white blood cell counts (see Supplementary Table~S2 for the full list of features). 
The missing data are irregularly distributed across the dataset, so we gather the attributes into 7 different clusters based on the overlap between the features in order to minimize the amount of information loss.
Figure~\ref{fig3}a shows the clusters labeled from $A$ to $G$ and the attributes included in each cluster. 
Figure \ref{fig3}b shows the input data color-coded based on the different clusters, from cluster $A$ in yellow to cluster $G$ in dark green. 
The colored areas represent the data, while the missing values are shown as white patches.
Discarding all patients with incomplete records leads to a reduced cohort of only $501$ subjects, as indicated by the black rectangle in Figure~\ref{fig3}b.

The schematic representation of the GapNet approach is provided in Figure~\ref{fig3}c.
We split the data into training and testing sets.
The testing set includes about $20\%$ ($N_{\rm test} = 100$) of the complete samples, while the training set includes the remaining complete data ($ 401 $ samples) and all the incomplete data ($N_{\rm train} = 3826$).

In training stage I, the seven clusters of features are used to train seven independent neural networks with input layer of 5 nodes, except for cluster A (with 4 nodes). Each of these neural networks has 2 hidden layers with 10 nodes each (ReLu activation) and a dropout layer (frequency 0.2). The output layer consist of a single neuron (sigmoidal activation). 
We train the neural networks on each cluster of inputs for 500 epochs using the binary cross-entropy loss function and Adam optimizer.
These seven networks are then combined into a single neural network in training stage II with a joint output node, and the new connections are retrained on the complete samples (500 epochs, binary cross-entropy loss function, Adam optimizer).

Figure~\ref{fig3}d shows that the ROC curve for the GapNet approach (orange, AUC $0.69 \pm 0.07$) is significantly better that that of the vanilla neural network (blue, AUC $0.60 \pm 0.07$) as demonstrated by the Delong test \cite{delong1988comparing} ($p<0.0001$, $z=11.8$).
The solid lines are the average ROC curves over $100$ repetitions of random splitting between the training and testing sets.
Table~\ref{tab1} shows that the sensitivity, specificity, accuracy, and precision computed by setting the predictions' threshold at $0.5$ are much improved by the GapNet approach.

Finally, we compare the performance of each cluster to GapNet and the vanilla neural network in Figure~\ref{fig3}e. 
The results are sorted in descending order of median AUC values. 
It is interesting to note that the GapNet results are better than those of the best cluster, demonstrating that combining the information in the clustered network improves the classification task. 
Moreover, the shift in the median between the GapNet and the subsequent best cluster is larger than the average distance between single clusters, showing a solid improvement in the predictions. 
Figure~\ref{fig3}e shows also that the training on some specific features (included in clusters $A$ and $E$) results in a better predictor than using all $34$ features while discarding the missing data. 
A possible explanation is that the vanilla neural network is trained on a smaller dataset (including only the records from $501$ patients), while the singular clusters have a larger size ($3875$ and $1992$ subjects for clusters $A$ and $E$, respectively). 

\section{Discussion}

The need to handle incomplete datasets is ubiquitous in all fields dealing with empirical data, such as medicine and engineering.
While several imputation techniques have been developed, they always rely on (explicit or implicit) assumptions on the frequency and distribution of missing values, and often incur the risk of introducing biases.
Here, we have proposed an alternative approach that can be employed also when the missing data are very frequent and not missing at random. We have called this approach GapNet.
In the GapNet approach, the neural network undergoes two training stages, first training an ensemble of neural networks, each on a data subset with a complete cluster of features, and then combining these into a single estimator that is fine-tuned on the available complete samples. 
This estimator is better at predicting the results for complete data when compared both with a simple vanilla neural network approach trained only on complete data and with the single-cluster-trained neural networks.
We have demonstrated the superior predictive ability of the GapNet approach on three examples, one corresponding to simulated data as a proof of principle, and the other two on real-word medical datasets of Alzheimer's disease and Covid-19 patients. These findings suggest that the improved predictions obtained by the GapNet approach are potentially 
generalizable to other datasets.

The major goal for the classification in the AD continuum is to establish a biomarker-based deep-learning model. 
In this respect, the GapNet approach outperforms the vanilla neural network, delivering robust results and detecting complex relationships when combining different imaging modalities. 
Noteworthy, the feature importance analysis shows that the important brain regions for the GapNet approach are in line with the brain areas affected in AD reported by previous studies.
Specifically, the most predictive features are mainly derived from the amyloid-PET modality, including parietal and frontal regions, which are typical sites of amyloid accumulation in AD \cite{Grothe2031}.
These key findings on AD-related brain changes suggest that the GapNet estimator is capable of producing robust predictions for early and accurate AD diagnosis.

A hot topic in COVID-19 research is to understand the effect of other comorbid diseases and conditions on the risk of developing severe Covid-19 symptoms \cite{foy2020ass, henry2020red, wang2020red,pakos2020characteristics, d2020coronavirus, hu2020declined,radenkovic2020cholesterol}.
This type of studies can help understanding, for instance, which individual's characteristics are associated with a higher risk of hospitalization, and who should be constantly monitored or prioritized in the vaccination process \cite{hassan2020should,cook2021impact, hezam2021covid}. 
However, when analyzing patient data, the choice between discarding the incomplete values or imputing them, and the imputation technique employed \cite{zhang2020novel} can lead to different results \cite{zeng2020can,bastug2020clinical, elliott2021covid} depending also on the size of the cohort and the handling of missing data.
Here, we provide a simple example to predict severe Covid-19 outcomes, with the aim of pointing out the advantages of using GapNet in this context: exploiting the incomplete values and avoiding biases or alterations of the original dataset.
In the presented results, the incidence of the different features is consistent with previous findings in the literature. 
The most relevant clusters, in fact, include features such as age, systolic and diastolic blood pressure and  serum creatinine levels (see details in Supplementary Note A, ``Feature importance analysis for Covid-19 severity''), connected with known Covid-19 high-risk comorbidities \cite{gallo2021predictors,lippi2020hypertension,d2020coronavirus}.

\section{Conclusion}

In conclusion, we have proposed GapNet, a conceptually effective model of neural network architectures, to produce more robust
predictions in datasets with missing values, which have become increasingly common in research.
We have shown how GapNet can detect complex nonlinear relationships between all the variables and is capable of learning and inferring from medical data with incomplete features.
We have verified the effectiveness of  GapNet in two real-world prominent datasets, the identification of patients in the Alzheimer’s disease continuum in the ADNI cohort, and the prediction of patients at risk of hospitalization due to Covid-19 in the UK BioBank cohort.
We believe that GapNet is a preliminary step towards generic, scalable architectures that can investigate many real-world medical problems, or even tasks from many domains, holding great potential for several future applications.
One of the next steps will be to apply GapNet to more complex kinds of neural network architectures, such as recurrent and convolutional neural networks, as well as to apply it to more complex input data, such as time sequences and images.

\section{Methods} \label{sec:method}

\subsection{Simulated dataset}

To verify the working principle of GapNet, we use a simulated dataset adapted from Madelon \cite{Guyon2005} implemented with scikit-learn \cite{pedregosa2011} (Figure~\ref{fig1}a).
We simulate a binary-classification dataset including $1000$ samples with $F=40$ features ($x_1, \dots, x_{40}$). 
Of these features, $25$ are informative features ($x_{1}$, $x_{3}$, $x_{4}$, $x_{5}$, $x_{8}$, $x_{10}$, $x_{11}$, $x_{13}$, $x_{14}$, $x_{15}$, $x_{18}$, $x_{20}$, $x_{22}$, $x_{25}$, $x_{26}$, $x_{27}$, $x_{28}$, $x_{29}$, $x_{31}$, $x_{33}$, $x_{34}$, $x_{35}$, $x_{36}$, $x_{39}$, $x_{40}$), $10$ are linear combinations of the informative features ($x_{2}$, $x_{6}$, $x_{7}$, $x_{9}$, $x_{12}$, $x_{17}$, $x_{19}$, $x_{24}$, $x_{30}$, $x_{37}$), and $5$ are uncorrelated random noise without information ($x_{16}$, $x_{21}$, $x_{23}$, $x_{32}$, $x_{38}$).
We introduce missingness in the dataset by removing the values of features $x_1$ to $x_{25}$ from samples $1$ to $450$ and the values of features $x_{25}$ to $x_{40}$ for samples $551$ to $1000$, so that only 100 samples (samples $451$ to $550$) are complete (Figure~\ref{fig1}b).

\subsection{ADNI cohort}

The data used for this analysis were obtained from the ADNI database (\url{adni.loni.usc.edu}).
The ADNI was launched in 2003 as a public--private partnership, led by Principal Investigator Michael W. Weiner, MD. 
The primary goal of ADNI has been to test whether serial magnetic resonance imaging (MRI), positron emission tomography (PET), other biological markers, and clinical and neuropsychological assessment can be combined to measure the progression of mild cognitive impairment (MCI) and early Alzheimer's disease (AD).
The individuals included in the current study were recruited as part of ADNI-GO and ADNI-2. 
This study was approved by the Institutional Review Boards of all participating institutions.
Informed written consent was obtained from all participants at each site.
For up-to-date information, see \url{adni.loni.usc.edu}.

Amyloid pathology has been identified using a cut-off of $>976.6\,{\rm pg/ml}$ on cerebrospinal fluid (CSF) levels of A$\beta$42, following previously established procedures \cite{hansson-2018}. Subjects who are cognitively normal and have high CSF A$\beta$42 values are used as a healthy reference group, while subjects who are cognitively normal, have mild cognitive impairment, or AD dementia with low CSF A$\beta$42 values are included in the group with high risk of having AD (the AD continuum group).

\subsection{UK Biobank cohort}

The dataset employed in the prediction of Covid-19 hospitalization is based on the general practitioners' records provided by the UK BioBank dedicated to Covid-related research \cite{ukbb}. 
We build the labels using the COVID-19 test result records and the hospital inpatient register data. 
The labels are assigned `0' for patients positive to Covid-19 who do not appear in the hospital records, and `1' for patients hospitalized due to Covid-19 and identified by the ICD-10 diagnostic code U07.1 \cite{WHO}. 
The input values come from merging the primary care data from the two largest providers, TPP \cite{TPP} and EMIS \cite{EMIS}. 
The data are cleaned to obtain the dataset: we select the patients positive to Covid-19, we filter only the medical coding systems SNOMED CT \cite{snomed} and CTV3 \cite{ctv3}, and we include records in the five years interval 1 January 2015 to 1 January 2020 (antecedent to the first reported case in the dataset).
The selected values include many standard medical examinations, such as pressure measurements, body weight, and blood tests.
At this point, we cut the less common codes to obtain a dataset of $501$ subjects with complete features and we undersample the more represented category (the non-hospitalized subjects are nearly five times larger) to obtain a balanced dataset.
The final number of attributes incorporated is $34$, listed in Table~\ref{stab2} together with their corresponding Read Code. 

\section{Acknowledgments}

We acknowledge support from the MSCA-ITN-ETN project \emph{ActiveMatter} sponsored by the European Commission (Horizon 2020, Project Number 812780). 

Data collection and sharing for this project was funded by the Alzheimer's Disease Neuroimaging Initiative (ADNI) (National Institutes of Health Grant U01 AG024904) and DOD ADNI (Department of Defense award number W81XWH-12-2-0012). ADNI is funded by the National Institute on Aging, the National Institute of Biomedical Imaging and Bioengineering, and through generous contributions from the following: AbbVie, Alzheimer’s Association; Alzheimer’s Drug Discovery Foundation; Araclon Biotech; BioClinica, Inc.; Biogen; Bristol-Myers Squibb Company; CereSpir, Inc.; Cogstate; Eisai Inc.; Elan Pharmaceuticals, Inc.; Eli Lilly and Company; EuroImmun; F. Hoffmann-La Roche Ltd and its affiliated company Genentech, Inc.; Fujirebio; GE Healthcare; IXICO Ltd.; Janssen Alzheimer Immunotherapy Research \& Development, LLC.; Johnson \& Johnson Pharmaceutical Research \& Development LLC.; Lumosity; Lundbeck; Merck \& Co., Inc.; Meso Scale Diagnostics, LLC.; NeuroRx Research; Neurotrack Technologies; Novartis Pharmaceuticals Corporation; Pfizer Inc.; Piramal Imaging; Servier; Takeda Pharmaceutical Company; and Transition Therapeutics. The Canadian Institutes of Health Research is providing funds to support ADNI clinical sites in Canada. Private sector contributions are facilitated by the Foundation for the National Institutes of Health (www.fnih.org). The grantee organization is the Northern California Institute for Research and Education, and the study is coordinated by the Alzheimer’s Therapeutic Research Institute at the University of Southern California. ADNI data are disseminated by the Laboratory for Neuro Imaging at the University of Southern California. 

This research has been conducted using data from UK Biobank, a major biomedical database, under the following application number: 37142. 


\end{document}